\documentclass[10pt,twocolumn,letterpaper]{article}

\usepackage{cvpr}
\usepackage{times}
\usepackage{epsfig}
\usepackage{graphicx}
\usepackage{amsmath}
\usepackage{amssymb}
\usepackage{caption}
\usepackage[pagebackref=true,breaklinks=true,letterpaper=true,colorlinks,bookmarks=false]{hyperref}

\cvprfinalcopy 

\def\cvprPaperID{****} 

\usepackage{fancyhdr}
\usepackage{lipsum} 

\fancypagestyle{plain}{
  \fancyhf{}
  \fancyfoot[C]{\thepage}%
}
\fancypagestyle{firstpage}{
  \fancyhf{}
  \fancyfoot[L]{* This work was done when the authors were at University of California, San Diego. 2021}%
}
\ifcvprfinal\fi
\begin{document}

\title{ Domain Adaptation for Object Detection using SE Adaptors and Center Loss}

\author{
Sushruth Nagesh*\\
UC San Diego\\
{\tt\small snagesh@ucsd.edu}
\and
Shreyas Rajesh* \\
UC San Diego\\
{\tt\small s1rajesh@ucsd.edu}
\and
Asfiya Baig*\\
UC San Diego\\
{\tt\small asbaig@ucsd.edu}
\and
Savitha Srinivasan*\\
UC San Diego\\
{\tt\small sasriniv@ucsd.edu}

}

\maketitle
\thispagestyle{firstpage}
\begin{abstract}
Despite growing interest in object detection, very few works address the extremely practical problem of cross-domain robustness especially for automative applications. In order to prevent drops in performance due to domain shift, we introduce an unsupervised domain adaptation method built on the foundation of faster-RCNN with two domain adaptation components addressing the shift at the instance and image levels respectively and apply a consistency regularization between them. We also introduce a family of adaptation layers that leverage the squeeze excitation mechanism called SE Adaptors to improve domain attention and thus improves performance without any prior requirement of knowledge of the new target domain. Finally, we incorporate a center loss in the instance and image level representations to improve the intra-class variance. We report all results with Cityscapes as our source domain and Foggy Cityscapes as the target domain outperforming previous baselines. 
\textcolor{red}{code: https://github.com/shreyasrajesh/DA-Object-Detection}
\end{abstract}

\section{Introduction}

Object detection is one of the early problems in computer vision with continued and diverse efforts being made to identify and localize all instances of various categories of objects in images from a variety of challenging datasets such as COCO \cite{coco}, KITTI \cite{kitti}, Cityscapes \cite{cityscapes} and more. Deep CNNs have proven extremely vital in this progress drastically improving performance, especially region-localization based approaches such as Faster-RCNN \cite{faster-rcnn}. However, most of these methods have been designed for performing well on a single domain/dataset and requires complete retraining on new and unseen labeled data which is extremely expensive. This low adaptability significantly reduces the impact of these methods, as they become unusable in real applications especially in automotive use cases. Small variations in lighting, weather conditions, time of day, location etc. is extremely common but requires entire retraining of models. Therefore, it is highly desirable to develop models that can adapt object detection to these new conditions (domains) that are visually different than the original domain.

\begin{figure*}[h!]
  \includegraphics[width=18.5cm,height=7.1cm]{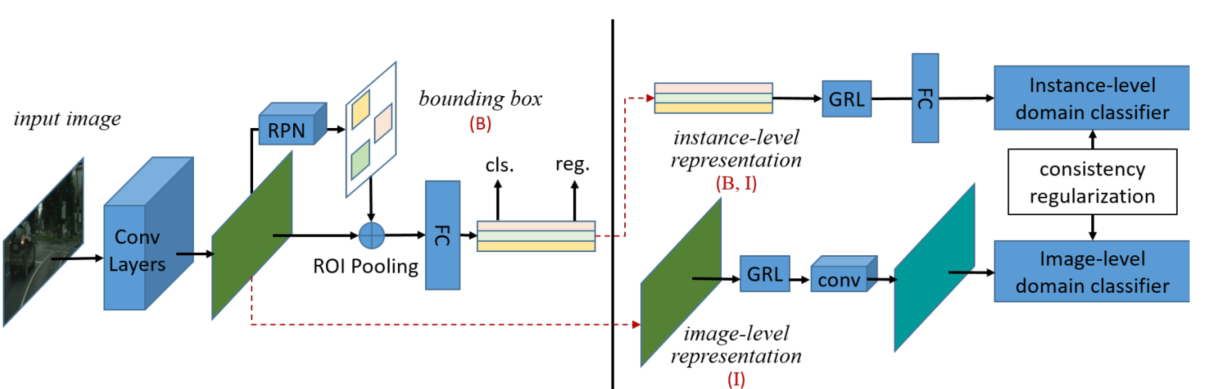}
  \caption{Baseline architecture: Domain-adaptive Faster-RCNN}
  \label{baseline}
\end{figure*}

In this work, we consider one such approach to perform unsupervised domain adaptation for cross-domain object detection. We use labeled data from one domain as the source data and unlabeled data from a new domain as our target data and aim to adapt our object detection model to perform well on both domains. In order to achieve this, we build an end-to-end domain adaptation model as shown in Figure \ref{baseline} by incorporating two domain adaptation components at the instance level and the image level into the Faster-RCNN \cite{faster-rcnn} architecture to reduce the $H$-divergence between the two domains. Each component learns a domain classifier and employs an adversarial strategy to learn domain-invariant features. We incorporate a consistency regularization loss, to learn domain-invariant region proposals. Further, we introduce Squeeze-Excitation Adaptors as introduced by Wang et. al, \cite{universal} for domain specific attention to improve our model and incorporate a center loss as proposed by Wen et al., \cite{center-loss} in each of the domain adaptation components to reduce the intra-class variance in the source and target domain feature space. We perform experiments and evaluate our results on the Cityscapes and Foggy Cityscapes datasets to demonstrate the superiority of our approach.

\section{Related Work}

\subsection{Object Detection}

 Among the large number of deep learning approaches, region-based CNNs have received significant attention due to their effectiveness. This line of work was pioneered by R-CNN \cite{rcnn}, which extracts region proposals from the image and a network is trained to classify each region of interest (ROI) independently. The two stage detection frameworks of Fast R-CNN \cite{fast-rcnn} and Faster R-CNN \cite{faster-rcnn} detectors have achieved great success in recent years, and laid the foundation for many follow up works such as R-FCN \cite{rfcn}, FPN \cite{fpn}, MS-CNN \cite{ms-cnn}. Additionally, single stage object detectors, such as YOLO \cite{yolo}, and SSD\cite{ssd} became popular for their fairly good performance and high speed. However, these works were only able to perform well on the domain they were trained on. In most practical situations, we face domain gaps between the train and test data. It is important to consider the domain adaptation issue for object detection across multiple datasets.

\subsection{Domain Adaptation}

Deep domain adpatation techniques have been successfully applied in many real-world applications, including image classification, and style translation. Conventional methods to address domain adaptation include asymmetric metric learning \cite{asym}, subspace alignment \cite{subspace-align}, covariance matrix alignment \cite{frustrating-easy} \cite{distance-min}, etc. However, fewer papers address adaptation beyond classification and recognition, such as object detection, face recognition, semantic segmentation and person re-identification. In this paper, we will focus on object detection across multiple domains.

\subsection{Domain Adaptation for Object Detection}
The recent works that address domain adaptation for object detection can be categorized based on the approaches used \cite{survey}. Discrepancy-based approaches \cite{discrepancy-od} diminish the domain shift by fine-tuning the deep network based detection model with labeled or unlabeled target data. Adversarial-based approaches \cite{wild} \cite{adversarial} utilize domain discriminators and conduct adversarial training to encourage domain confusion between the source domain and the target domain. Reconstruction-based approaches \cite{recon} presume that the reconstruction of the source or target samples is helpful to improve the performance of domain adaptation object detection. We will build upon the work of Chen et al \cite{wild} who use adversarial learning to address object detection in the wild. 

\subsection{Domain Attention}
In \cite{universal}, Wang et al build a universal object detection system that is capable of working on various image domains. Unlike multi-domain models, this universal model does not require prior knowledge of the domain of interest. They introduce a new family of adaptation layers, based on the principles of squeeze and excitation \cite{senet}, and a new domain-attention mechanism. A Universal SE bank captures the feature subspaces of the domains spanned by all datasets, and the attention \cite{attention} mechanism soft-routes the USE projections. We incorporate the domain attention mechanism to improve the adaptation task. 

\subsection{Center Loss}
In \cite{center-loss}, Wen et al propose a new loss function called center loss to efficiently enhance the discriminative power of the deeply learned features in neural networks. Specifically, a center is learned (a vector with the same dimension as a feature) for deep features of each class. In our paper, we add the center loss to the domain classifier loss.

\section{Methodology}
We use the domain-adaptive Faster-RCNN model proposed in \cite{wild} as our baseline. The paper introduces two domain adaptation components into the Faster-RCNN model to align the feature distributions at the image and instance levels. Additionally, we also introduce a universal Squeeze and Excitation (SE) adaptor bank into this baseline model. We also add a center loss component to the image and instance-level classifiers to achieve a tighter bound on the intra-class variances. The center loss is back propagated only till the gradient reversal layer. 

\subsection{Image-level adaptation}
A patch-based domain classifier is used to eliminate domain distribution mismatch at the image level. The domain classifier outputs a domain label for each patch in the image. The idea is to address the shift at a global level such as image style, illumination, scale etc. 

\subsection{Instance-level adaptation}
The instance-level adaptation is performed on the ROI-based feature vectors before feeding it into the last fully-connected layer. This is done to address shift at a local level such as object appearance, viewpoint etc. A domain classifier is trained to align the feature vectors at the instance level as well. A gradient reversal layer is added before the domain classifier to incorporate an adversarial training strategy. 

\subsection{$\mathcal{H}$-Divergence}
The domain adaptation components at the image and instance level are used to reduce the $\mathcal{H}$-Divergence between the two domains. The $\mathcal{H}$-Divergence defines the distance between the two domains as: 
\begin{equation}
    d_H(S,T) = 2 \Big(1 - min_{h \in H} \big( err_s(h(\textbf{x})) + err_T(h(\textbf{x}))\big)\Big)
\end{equation}
where $\textbf{x}$ is a feature vector, $h: \textbf{x}\rightarrow \{0,1\}$ is a domain classifier, and $err_S$ and $err_T$ are the prediction errors of $h(\textbf{x})$ on the source and target domain samples, respectively. 
To align the two domains, we therefore need to minimize $d_H(S,T)$:
\begin{equation}
    \underset{f}{\min} \;{d_H(S,T) \Leftrightarrow \underset{f}{\max} \underset{h \in H}{\min} \{err_S(h(\textbf{x})) + err_T(h(\textbf{x})\}}
\end{equation}

This objective is optimized using adversarial training.

\subsection{Domain Attention}
We adopt the domain attention module proposed in \cite{universal} in order to address non-trivial domain shifts. The domain attention module consists of a universal Squeeze and Excitation (SE) adaptor bank. The adaptor bank is introduced into a few convolutional layers of the feature extractor and the Region Proposal Network as a feature-based attention mechanism. The SE adaptor bank enables the module to learn activations specific to domains. 
\begin{figure*}[h!]
\includegraphics[width=185mm, height=73mm]{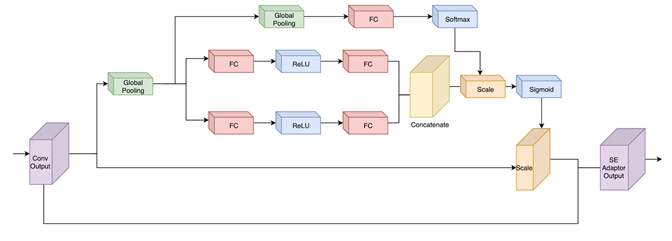}
\caption{SE Net adaptor block}
\label{senet}
\end{figure*}

\subsubsection{SE Adaptor bank}
SE Networks \cite{senet} improve channel inter dependencies at almost no additional computational cost. It is a 'content aware' mechanism that assigns weights to different channels when creating output feature maps. An SE adaptor block, as shown in fig \ref{senet}, consists of the following components: 1. A global pooling layer (that squeezes each channel in the input), 2. A fully connected layer followed by ReLU that adds necessary non linearity, 3. A second fully connected layer followed by Sigmoid activation. The SE block outputs weights for the different channels in the input. The computation performed by the SE network is thus given by: 

\begin{equation}
    \textbf{X}_{SE} = \textbf{F}_{SE}(\textbf{F}_{avg}(\textbf{X}))
\end{equation}
where $ \textbf{F}_{avg} $ is a global average pooling operator, and $\textbf{F}_{SE}$ is the combination of FC1 with ReLU activation and FC2 with Sigmoid activation. As proposed in \cite{universal}, we generalize the SE block to an adaptor bank, as shown in fig \ref{senet}.

\subsection{Image and Instance level Center loss}
The main aim of using center loss is to decrease the intra class variance of both source and target domain feature space. This can be implemented by adding center loss to the domain classifier loss. The domain classifier has a min max objective function as it follows an adversarial approach. This characteristic is desired for the softmax loss function which tries to separate the two domains in the domain adaptation layers and bring them together in the backbone network using GRL layer.

On the other hand, it is always desirable to have less intra class variance be it the domain adaptation layers or the backbone layers. This calls for minimizing the center loss all together. In our simplified approach, we minimize the center loss only in the domain adaptation layers. We mask the gradients due to center loss in the backbone network.

Center loss is employed at both image level and instance level domain adaptation layers. At both places we initialize two centers of appropriate dimensions representing source and target domains respectively. The image level center loss can be represented as below
\begin{equation}
L_{img\_c} = \frac{1}{2} \sum_{i=1}^{m} \parallel x_{i} - c_{img\_y_{i}}\parallel_{2}^{2}
\end{equation}
where $c_{img\_y_{i}} \epsilon R^{dimg}$ denotes the $y_{i}$th class center. $m$ denotes the batch size of the current iteration. For the image level domain adaptation we try to classify an image as a whole.

In the instance level domain adaptation component, since we are trying to classify every rpn proposal, each image will have $n$ samples to classify. Here $n$ denotes the number of anchor box proposals. The instance level center loss can be represented as below
\begin{equation}
    L_{inst\_c} = \frac{1}{2} \sum_{i=1}^{n*m} \parallel x_{i} - c_{inst\_y_{i}}\parallel_{2}^{2}
\end{equation}
$m$ denotes the batch size of the current iteration. $c_{inst\_y_{i}} \epsilon R^{dinst}$ denotes the $y_{i}$th class center.

The masking effect for this loss is achieved by employing a separate optimizer spanning only the weights of the domain adaptation layers. This optimizer is used with the main optimizer synchronously. Gradients are estimated for both the optimizers before back-propagating any of it. 

\subsection{Instance and Image level Cross Entropy loss}

For the domain classifier, cross entropy loss as proposed in \cite{wild} is used. The image level cross entropy loss can we written as,
\begin{equation}
    L_{ce} = \sum _{i,u,v} D_{i} \log p_{i}^{u,v} + (1-D_{i}) \log(1-p_{i}^{u,v})
\end{equation}
where $D_i = 0$ represents source domain and $D_i = 1$ represents target domain. Here each pixel of the domain classifier output is classified. So, $(u,v)$ spans all the positions of the output feature map of the domain classifier.

The instance level cross entropy loss can be written as,
\begin{equation}
        L_{ce} = \sum _{i,j} D_{i} \log p_{i,j} + (1-D_{i}) \log(1-p_{i,j})
\end{equation}

where output of instance level domain classifier for the j-th domain in the i-th image is $p_{i,j}$

\subsection{Consistency loss}

As in \cite{wild}, we need to enforce consistency between image and instance level domain classifiers. One way to do that is to enforce both the classifiers to give similar class of prediction for a given image. Since image level domain classifier produces pixel wise output, we consider average of all these outputs as the image level probability. The consistency loss can be written as,
\begin{equation}
    L_{cst} = \sum_{i,j} \parallel\frac{1}{I} \sum_{u,v} p_{i}^{u,v} - p_{i,j}\parallel_{2} 
\end{equation}
where $|I|$ represents total number of pixel predictions per image

The total loss expression in two steps can be represented as below
\begin{equation}
    L_{1} = L_{sup} + \lambda_{ce} (L_{img\_ce} + L_{inst\_ce} + L_{cst})
\end{equation}
The center losses can be written as,
\begin{equation}
    L_{2} = \lambda_{\text{center}}(L_{img\_c} + L_{inst\_c})
\end{equation}

where $\lambda_{ce}$ and $\lambda_{\text{center}}$ are the weights of the cross entropy and center losses respectively.

\section{Experimental Setup}

\subsection{Datasets}
We use Cityscapes \cite{cityscapes} as our source domain, with images dominantly obtained in clear weather. For the target domain, we use the Foggy Cityscapes dataset. It is a synthetic foggy dataset in that it simulates fog on real scenes. The images are rendered using the images and depth maps from Cityscapes. The semantic annotations and data split of Foggy Cityscapes are inherited from Cityscapes, making it ideal to study the domain shift caused by weather condition.

\subsection{Training Details}
The code is developed using PyTorch framework. The Faster R-CNN network architecture employed uses a ResNet 50 backbone. GRL layer proposed in \cite{GRL} is used to reverse the gradients for adversarial training. The model is initialized with weights trained on ImageNet.     

Unless mentioned the following setting is used for both the optimizers proposed. Initially, lr = 0.001 is used for 50K iterations. Then it is decayed to 0.0001 for the next 20k iterations. Also momentum=0.9 is used for all iterations. Batch size = 2 is used for training. 

256 anchor boxes per image of sizes (32, 64, 128, 256, 512) and aspect ratios (1.0,2.0,3.0) are used. For all experiments, NMS threshold = 0.3 and confidence threshold = 0.8 are used. The weights for the image level and instance level domain classifier cross entropy losses are fixed to 0.1. Experiments to find optimal weights for center losses are carried out. COCO format mAP values are reported which use IOU from 0.5 to 0.95 in steps of 0.05.

\section{Results \& Discussion}

The baseline is compared against 3 models. SE+DA-FRCNN is trained by incorporating domain attention modules in stage 3 of the ResNet50 backbone of DA-Faster-RCNN. Center+DA-FRCNN uses the center loss as a domain classifier loss. SE+Center+DA-FRCNN contains domain attention modules in stage 3 of the backbone, and also uses the center loss for domain confusion. The quantitative results can be seen in table \ref{Overall}. We observe that individual use of domain attention and center loss have resulted in improvements in the mAP by a small margin. However, when domain attention and center loss are both used to train a model, there is a drop in performance.  

\begin{table}
\begin{center}
\begin{tabular}{ |c|c|c|c|c| } 
\hline
Model & mAP \\
\hline
DA-Faster-RCNN & 24.9 \\
\hline
SE + DA-FRCNN & 25.11 \\
\hline
Center + DA-FRCNN & 25.01 \\
\hline
SE + Center + DA-FRCNN & 24.3 \\
\hline
\end{tabular}
\end{center}
\caption{Quantitative results on the Foggy Cityscapes validation set, models are trained on the Cityscapes training set}
\label{Overall}
\end{table}

\begin{figure*}[htp!]
\begin{center}
\includegraphics[width=19cm, height=15cm ]{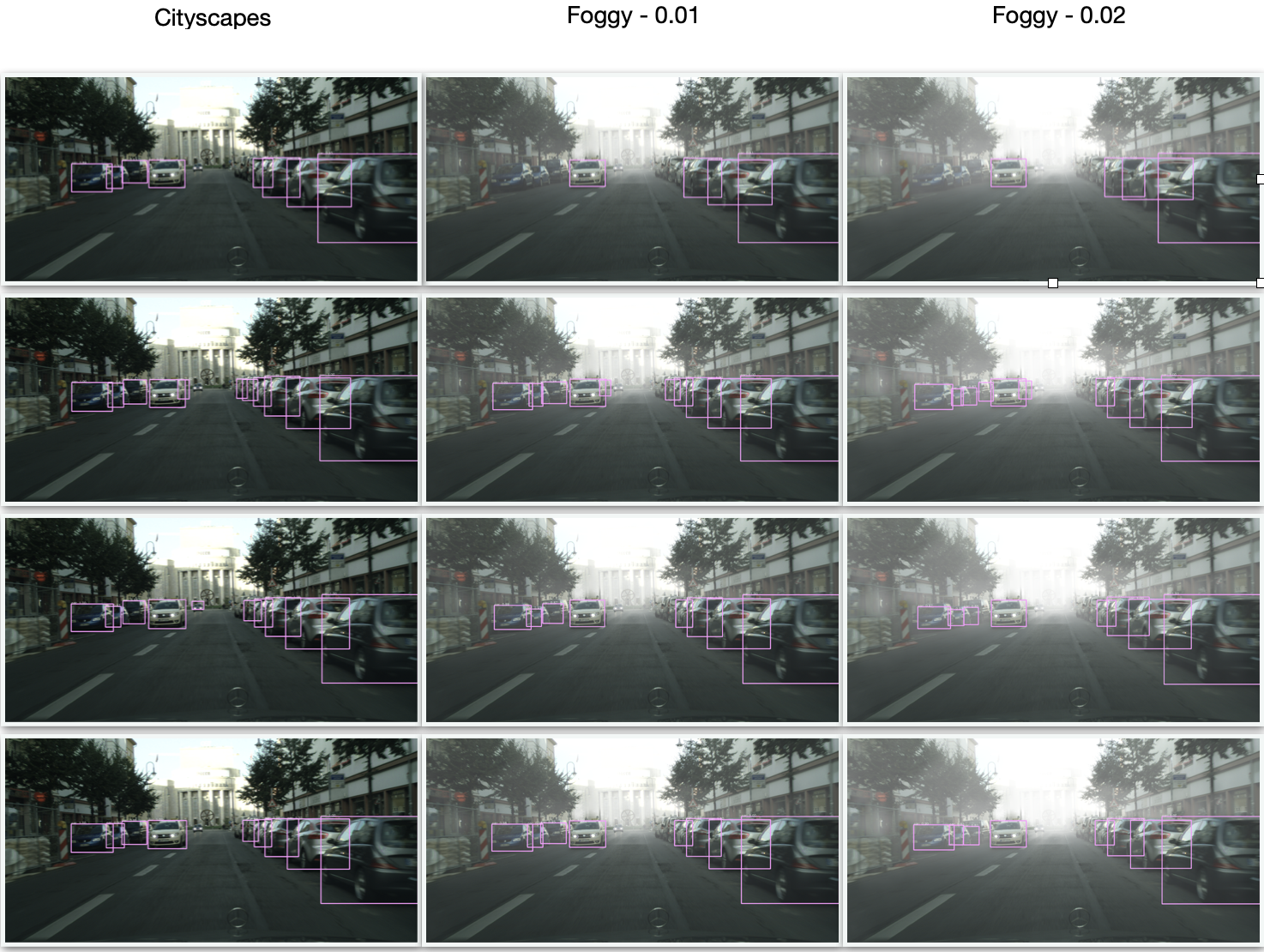}
\caption{Cityscapes images comprise the source domain, whereas Foggy cityscapes images comprise the target. Four rows corresponding to the four results reported in Table \ref{Overall}. Top to Bottom: DA Faster-RCNN, SE with DA FRCNN, Center Loss with DA FRCNN and SE + Center Loss with DA FRCNN.}
\end{center}
\label{results}
\end{figure*}

We compare the qualitative results of the 4 models in Figure \ref{results}. We see that SE+DA-FRCNN gives us more confident predictions and improved detection performance overall compared to the baseline. We also note that center loss improves the detection performance when used independently. However, in the qualitative results of SE+Center+DA-FRCNN, we note a dip in performance. 

\subsection{Ablation study: Effect of USE Bank}
Two experiments were conducted to analyze the effect of using the USE banks in the ResNet50 backbone. In the first experiment, USE banks are used in all 3 stages of the backbone. In the second experiment, USE banks are used only in the 3rd stage of the backbone. The results, as shown in table \ref{USE}, are compared against the baseline which does not contain any USE banks in the ResNet50 backbone. It is observed that having USE banks in all stages, as in experiment 1, causes a drop in the performance. This could be because the early stages of the backbone contain low-level features like edges, which are not domain dependent. Therefore, there is no need for domain attention in stages 1 and 2 as they constitute the early layers of a CNN. For experiment 2, where the USE banks are present in stage 3, we observe an improvement compared to the baseline. These results give us 2 key insights. Firstly, using domain attention in early stages may cause a drop in performance. Secondly, applying domain attention in later stages containing high-level features can result in improvements in performance.  

\begin{table}
\begin{center}
\begin{tabular}{ |c|c|c|c|c| } 
\hline
 & Stage 1 & Stage 2 & Stage 3 & mAP \\
\hline
DA-Faster-RCNN &  &  & & 24.9 \\
\hline
Ours & \checkmark & \checkmark & \checkmark & 24.4 \\ 
 \hline
Ours & &  & \checkmark & \textbf{25.11} \\ 
\hline
\end{tabular}
\end{center}
\caption{Effect of USE banks in different stages of the Resnet50 backbone}
\label{USE}
\end{table}

\subsection{Ablation study: Effect of Center Loss}
To study the effect of center loss, two experiments were conducted. One experiment with more emphasis on center loss by having center loss weight $\lambda_{\text{center}} >> \lambda_{\text{ce}}$ was conducted. Here $\lambda_{\text{center}} = 1.0$ was used. Another experiment with a moderate center loss weight to balance the effect of both losses was conducted. Here $\lambda_{\text{center}} = 0.5$ was used.

It was observed that for both models, the mAP values increased compared to the baseline model. This improved performance can be seen in Table \ref{ablation}. Adding more weight to the center loss is clearly improving the mAP values. This justifies the theory behind using center loss for the domain adaptation layers. By decreasing the intra class variance, the number of very hard / noisy negatives is being decreased. This in turn helps in generating better training signal for the overall network.

\begin{table}
\begin{center}
\begin{tabular}{ |c|c|c|c|c| } 
\hline
Model & mAP \\
\hline
Center ($\lambda_{center} = 1.0)$ + DA-FRCNN & 25.01 \\
\hline
Center ($\lambda_{center} = 0.5)$ + DA-FRCNN & 24.94 \\
\hline
\end{tabular}
\end{center}
\caption{Effect of Center loss with different weights}
\label{ablation}
\end{table}

\section{Conclusion}

From our experiments, we can conclude that adding USE banks and Center loss is beneficial for the domain adapted object detection model. Adding USE adapter banks allows the model to be more domain attentive. It allows the network to switch between sub branches according to the domain of the input image. Using Center loss aids in generating better training signal by reducing the effect of noisy / hard negatives. But USE banks and Center loss do not seem to complement each other when used together. This can be due to improper weights for the center losses. Also it is possible that domain alignment is not desirable for SE adaptors because they maintain separate branches for different domains and center loss is just trying to aid domain alignment. Future work involves analyzing the combined effect of USE banks and Center loss to find if they indeed complement each other or not. 

{\small
\bibliographystyle{ieee}
\bibliography{egbib}
}

\end{document}